\PassOptionsToPackage{table,dvipsnames}{xcolor}
\documentclass[]{fairmeta}

\usepackage{hyperref}       
\usepackage{url}            
\usepackage{booktabs}       
\usepackage{amsfonts}       
\usepackage{nicefrac}       
\usepackage{microtype}      
\usepackage[normalem]{ulem}
\usepackage{array}
\usepackage{ltablex}
\usepackage{tabularx}
\usepackage{tabulary,multirow,overpic}
\usepackage{lscape}
\usepackage{makecell}
\usepackage{multirow}
\usepackage{amsmath}
\usepackage[most]{tcolorbox}
\usepackage{lipsum} 
\usepackage{courier} 
\usepackage[ruled,linesnumbered]{algorithm2e}
\usepackage{graphicx}
\usepackage{listings}
\usepackage{longtable}
\usepackage{todonotes}
\usepackage{rotating}
\usepackage{multicol}
\usepackage{multirow}
\usepackage{pifont}
\usepackage{wrapfig}
\usepackage{pgf}

\newcolumntype{Y}{>{\raggedright\arraybackslash}X}
\newcolumntype{Z}{>{\centering\arraybackslash}m{10mm}}

\newcommand{\g}[1]{\textcolor{gray}{#1}}
\definecolor{lightpurple}{RGB}{230, 220, 250}

\definecolor{borderpurple}{RGB}{120,81,169}
\definecolor{titlepurple}{RGB}{88,41,136}

\definecolor{lightgreen}{HTML}{D8ECD1}

\title{Meta CLIP~2: A Worldwide Scaling Recipe}

\author[1,2,*]{\color{borderpurple}{Yung-Sung Chuang}}
\author[1]{\color{borderpurple}{Yang Li}}
\author[1]{\color{borderpurple}{Dong Wang}}
\author[1]{Ching-Feng Yeh}
\author[1]{Kehan Lyu}
\author[1]{Ramya Raghavendra}
\author[2]{James Glass}
\author[1]{Lifei Huang}
\author[1]{Jason Weston}
\author[1]{Luke Zettlemoyer}
\author[1,\$]{Xinlei Chen}
\author[3]{Zhuang Liu}
\author[4]{\\Saining Xie}
\author[1]{Wen-tau Yih}
\author[1,\dagger]{\color{borderpurple}{Shang-Wen Li}}
\author[1,\dagger]{\color{borderpurple}{Hu Xu}}

\affiliation[1]{FAIR, Meta}
\affiliation[2]{MIT}
\affiliation[3]{Princeton University}
\affiliation[4]{New York University}

\contribution[]{\color{borderpurple}{\textbf{Core Contributors}}}
\contribution[*]{Work done during an internship}
\contribution[\$]{Work done when working at Meta}
\contribution[\dagger]{Project Leads}

\abstract{Contrastive Language-Image Pretraining (CLIP) is a popular foundation model, supporting from zero-shot classification, retrieval to encoders for multimodal large language models (MLLMs). 
Although CLIP is successfully trained on billion-scale image-text pairs from the English world, scaling CLIP's training further to learning from the worldwide web data is still challenging:
(1) no curation method is available to handle data points from non-English world;
(2) the English performance from existing multilingual CLIP is worse than its English-only counterpart, i.e., ``curse of multilinguality'' that is common in LLMs.
Here, we present Meta CLIP 2, the first recipe training CLIP from scratch on worldwide web-scale image-text pairs.
To generalize our findings, we conduct rigorous ablations with minimal changes that are necessary to address the above challenges and present a recipe enabling mutual benefits from English and non-English world data.
In zero-shot ImageNet classification, Meta CLIP 2 ViT-H/14 
surpasses its English-only counterpart by 0.8\%
and mSigLIP by 0.7\%,
and surprisingly sets new state-of-the-art without system-level confounding factors (e.g., translation, bespoke architecture changes) on multilingual benchmarks, such as CVQA with 57.4\%, Babel-ImageNet with 50.2\% and XM3600 with 64.3\% on image-to-text retrieval.
}

\date{\today}
\correspondence{Hu Xu \email{huxu@meta.com}, Shang-Wen Li \email{shangwel@meta.com}}

\metadata[Code and Model]{\url{https://github.com/facebookresearch/MetaCLIP}}

\begin{document}

\maketitle

\section{Introduction}
\label{intro}

Contrastive Language-Image Pre-training (CLIP)~\citep{radford2021learning} has become an essential building block of modern vision and multimodal models, from zero-shot image classification and retrieval to serving as vision encoders in multimodal large language models (MLLMs)~\citep{grattafiori2024llama,team2023gemini,liu2023visual,bai2023qwenvl}. 
CLIP and its majority variants \citep{ilharco_gabriel_2021_5143773,xu2024demystifying} adopt an English-only setting, and
Meta CLIP~\citep{xu2024demystifying} introduces a scalable data curation algorithm to meticulously extract a billion-scale English dataset that exhausts long-tailed concepts in Common Crawl.
The algorithm transforms the distribution of the raw Internet into \textit{controllable} and \textit{balanced} training distribution defined by metadata (e.g., visual concepts composed by human experts) and training distribution is known as one key contributor to performance.
In contrast, popular CLIP reproductions outsource such key contributor to external resources, e.g., OpenCLIP~\citep{ilharco_gabriel_2021_5143773} trained on LAION~\citep{schuhmann2021laion, schuhmann2022laion} and DFN~\citep{fang2023data} rely on pretrained CLIP models for black-box \emph{filtering} to keep only high-confidence data. Such approaches resemble distillation of an existing CLIP teacher model and produce \textit{untractable} distributions owned by an outsourcing party.

\begin{figure}[ht!]
    \centering
    \includegraphics[width=1.0\textwidth]{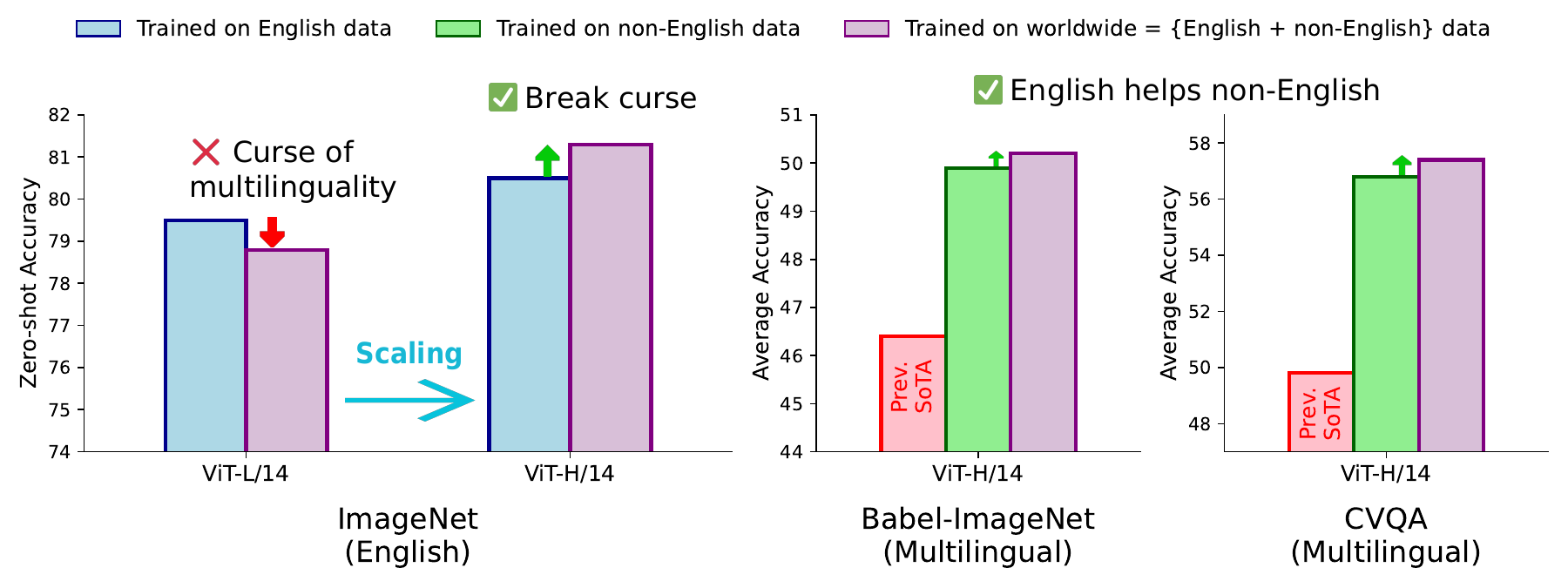}
    \caption{(Left) CLIP training suffers from the \textit{curse of multilinguality} that the English performance of a CLIP model trained on worldwide (i.e., English + non-English), billion-scale data is worse than its English-only counterpart, even when applying our recipe on ViT-L/14;
    scaling to ViT-H/14 enables non-English data helps English-only CLIP. (Right) English data also helps non-English CLIP.
    }
    \label{fig:teaser}
    \vspace{-2.mm}
\end{figure}

Although being the most widely used ``foundation'' models, most CLIP variants, including the scalable Meta CLIP, rely on English-only curation and thus discard the other, e.g., 50.9\%~\citep{internetLangStats2025} of non-English, worldwide web data. To extend CLIP training and data to the worldwide web for the next level of scaling, we inevitably have to handle these non-English image-text pairs—a barrier we refer to as the \textit{worldwide scaling challenges}, which are issues not yet being solved after years of attempts to train CLIP on multilingual data:
\newpage
\textbf{Challenge \#1: Lack of a fundamental data curation method to handle non-English data at scale}. Existing attempts either conduct no curation on the raw, non-English image-text pair data at all (e.g., distilling from English CLIP~\citep{chen2023mclip} or machine translation~\citep{carlsson2022cross,nguyen2024multilingual}), or rely on proprietary and private data sources (e.g., WebLI~\citep{chen2023pali} that drives mSigLIP and SigLIP 2~\citep{zhai2023sigmoid, tschannen2025siglip} is built from Google Image Search~\citep{juan2019graph}).

\textbf{Challenge \#2: Worse English performance than English-only CLIP}. This is also known as \textit{curse of multilinguality} in text-only large language models (LLMs).
For instance, mSigLIP is \(1.5\)\% worse than its English-only counterpart, SigLIP, on ImageNet~\citep{zhai2023sigmoid}, while SigLIP~2~\citep{tschannen2025siglip} prioritizes English performance at the cost of even worse multilingual results than mSigLIP. Hence, disparate models have to be used to optimize English and non-English performance at the same time.

\paragraph{This work.} We present \textbf{Meta CLIP~2}, the first ever recipe developing CLIP with training \emph{from scratch} on \emph{native} \textbf{worldwide} image-text pairs, without relying on outsourced resources, such as any private data, machine translation, or distillation. We empirically show that the curse of multilinguality in CLIP is the consequence of \emph{insufficient scaling} due to the lack of a proper recipe for worldwide data curation and model training.
When metadata, data curation, model capacity, and training are carefully designed and scaled \emph{jointly}, we show that not only the performance trade-offs between English and non-English data disappear, but the two become \emph{mutually beneficial}. Achieving such worldwide scaling is highly desirable, especially when English Internet data is exhausted soon~\citep{villalobos2022will}.

Our Meta CLIP~2 recipe is built on top of English Meta CLIP, where overlapping with OpenAI CLIP's vanilla architecture is deliberately maximized. The overlap makes our findings generalizable to CLIP and its variants, compared to system works (c.f.,~\citep{zhai2023sigmoid, tschannen2025siglip, bolya2025perception}) aiming at state-of-the-art (SoTA) performance with combination of all available techniques. Such combination involves confounding factors or comparison on outsourced resources instead of CLIP itself.
The Meta CLIP~2 recipe introduces three principled innovations for scaling to worldwide. 
1) \textit{Metadata.}  We scale the English Meta CLIP metadata 
to \(300\)+ languages on Wikipedia and multilingual WordNet.
2) \textit{Curation algorithm.} We build per-language substring matching and balancing to curate concept distribution for non-English data similar to the English counterpart.
3) \textit{Training framework.}
We design the first worldwide CLIP training framework, including an increase of seen image-text pairs during training proportional to the increased data size from the added non-English data examples, and a study on minimal viable model capacity to learn from worldwide scale data. 
As shown in Fig.~\ref{fig:teaser}, although a ViT-L/14 (the largest model size used by OpenAI) still suffers the curse of multilinguality, ViT-H/14 breaks the curse. English accuracy \emph{rises} from 80.5\% to 81.3\% on ImageNet and surprisingly new SoTA is set with minimal CLIP architecture changes for multilingual image-to-text retrieval (XM3600 64.3\%, Babel-ImageNet 50.2\%, and CVQA 57.4\%).

Together, Meta CLIP~2 enables the following desirable results by nature.
1) \textbf{Mutual benefits from the English and non-English worlds.} Non-English data now can better support an English-only model and vice versa, which is critical in the era when English data is depleting. 
2) \textbf{Full multilingual support.} Meta CLIP~2 never drops image-text pairs simply by languages and yields models outperforming all the previous multilingual systems, such as mSigLIP~\citep{zhai2023sigmoid} and SigLIP~2 \citep{tschannen2025siglip}.
3) \textbf{Native-language supervision.} Models learn directly from
alt-texts written by native speakers rather than synthetic machine
translations~\citep{pouget2024no, nguyen2024multilingual}.
4) \textbf{Cultural diversity.} Meta CLIP~2 retains the entire global distribution of images and thus inherits the comprehensive cultural and socioeconomic coverage advocated by \citep{pouget2024no}. Such coverage improves geo-localization and region-specific recognition.
5) \textbf{No-filter philosophy.} 
With the curation algorithm designed towards worldwide data, Meta CLIP~2 removes the last filter (i.e., whether the alt-text is in English) in pipeline, achieving better diversity and minimizing biases introduced by filters~\citep{pouget2024no}.
6) \textbf{Broader impacts on foundation data.} This work provides a foundational data algorithm designed for worldwide scale, and benefits not only CLIP, but also efforts using CLIP data such as MLLM~\citep{grattafiori2024llama,team2024chameleon}, SSL (Web-DINO~\citep{fan2025scaling}) and image generation (DALL-E~\citep{ramesh2021zero} and diffusion models~\citep{yang2023diffusurvey}).

\section{Related Work}
\label{related_work}

\subsection{Evolution of CLIP and its Data Processing}
CLIP~\citep{radford2021learning} and its variants~\citep{jia2021scaling,ilharco_gabriel_2021_5143773,zhai2023sigmoid} learn versatile image and text representations that are generally useful for downstream tasks~\citep{grattafiori2024llama,dai2023instructblip,liu2023visual}. Such multimodal contrastive learning and transformer architectures become standard components in vision and multimodal research.
Data is a key contributor to CLIP's performance~\citep{gadre2023datacomp,xu2024demystifying}. 
Two major processing approaches for CLIP data emerge: curation\footnote{Here, ``curation'' refers to select and align training data distribution with human from raw data source, excluding data filtering that is also referred to as curation in many works like DataComp~\citep{gadre2023datacomp, li2024datacomp} and DFN~\citep{fang2023data}.} from scratch, and distillation from external resources.
One key difference is that the former yields more \textit{controllable distribution} and the latter has untractable distribution owned by an outsourcing party.

\textbf{Curation from scratch.}
OpenAI CLIP~\citep{radford2021learning} curates a training dataset of 400M image-text pairs from scratch and publicizes high-level curation guidance. Meta CLIP~\citep{xu2024demystifying} makes OpenAI's guidance as a formal curation algorithm and scales the curation to 2.5B pairs.
The algorithm is model-free, no blackbox filtering, and fully transparent to enable training entirely from scratch on public data source, where the data distribution is curated to align with metadata composed by human experts (e.g., WordNet and Wikipedia). %

\textbf{Distillation from external resources}.
Distillation-based methods usually have good performance and save compute by learning from teacher model's knowledge~\citep{hinton2015distillingknowledgeneuralnetwork}.
However, in the context of CLIP training the teacher is usually an external blackbox system, which introduces untractable bias. 
For example, LAION-400M/5B~\citep{schuhmann2021laion,schuhmannlaion} (used by OpenCLIP~\citep{ilharco_gabriel_2021_5143773}) relies on OpenAI CLIP-filter and DFN~\citep{fang2023data} using a filter model trained on high-quality private data~\citep{ranasinghe2023perceptual}.
Recently, SigLIP~\citep{zhai2023sigmoid} and SigLIP~2~\citep{tschannen2025siglip}
learn from data source WebLI~\citep{chen2023pali}, which is derived
from Google Image Search~\citep{juan2019graph}.

\subsection{Vision Encoding}
\label{sec:related_ve}

CLIP-style models are widely used as vision encoders in MLLM, where language supervision in CLIP training helps to learn compact and semantic-rich visual representations. In contrast, traditional visual representation learning is based on self-supervised learning (SSL) methods like SimCLR~\citep{chen2020simple}, DINOv2~\citep{oquab2024dinov2}, and purely relies on the full visual signal without language bias.
There are variants that take advantage of both. 
SLIP~\citep{mu2021slip} combines language and SSL supervision; LiT~\citep{zhai2022lit} trains a vision encoder first and conducts language alignment later; Perception Encoder~\citep{bolya2025perception} shows early layers of CLIP representation yields vision-driven features with less semantic alignment. Recently, Web-DINO~\citep{fan2025scaling} shows SSL has better scalability on Meta CLIP curated large-scale data.
In summary, CLIP focuses on human-aligned representations optimized for compact models and efficient downstream uses; SSL models aim to preserve all visual information as a 
general pretraining approach. We envision more synergy from the two research lines due to complementarity.

\subsection{Multilingual CLIP Models}
Due to the lack of open source curation for public worldwide data, initial attempts to multilingual CLIP models are mainly distillation approaches. M-CLIP~\citep{carlsson2022cross} and mCLIP~\citep{chen2023mclip} simply leverage existing English-only CLIP as the vision encoder and trains a multilingual text encoder with low-quality multilingual pairs. To incorporate non-English data, subsequent works~\citep{santos2023capivara, nguyen2024multilingual, pouget2024no} leverage machine translation techniques, either translating non-English captions into English or vice versa. These distillation-based models carry existing English CLIP bias or translation bias on nonhuman-captioned data.
mSigLIP~\citep{zhai2023sigmoid} substantially advanced multilingual performance
by leveraging multilingual data from WebLI~\citep{chen2023pali}, which is an undisclosed dataset built with private data processing pipeline instead of publicly available worldwide data curation algorithm.

However, mSigLIP and other multilingual CLIP models suffer from the curse of multilinguality, e.g., mSigLIP is 1.5\% worse in ImageNet accuracy than its English-only counterpart SigLIP. Recently, SigLIP~2 adopts a notably English-centric design of having 90\% of its data in English, which is much higher than mSigLIP.
Mixed results are also observed ~\citep{wang2025scaling} on English benchmarks when scaling SigLIP from WebLI's 10B to 100B raw data, suggesting the challenges of scaling WebLI beyond.

\section{The Meta CLIP~2 Recipe}
\label{curation}

\begin{figure}[h]
    \centering
\includegraphics[width=0.95\textwidth]{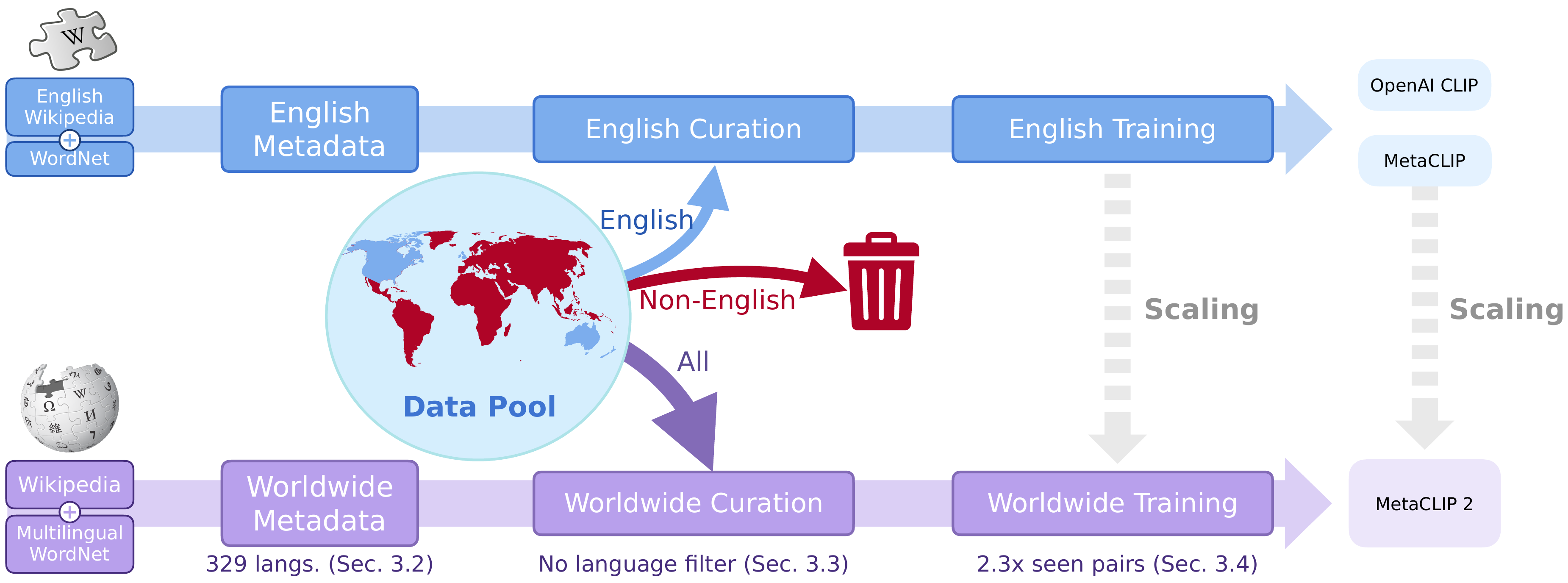}
    \caption{Overview of Meta CLIP 2 recipe: scaling CLIP data and training to worldwide scope.}
    \label{fig:pipeline}
    \vspace{-2.mm}
\end{figure}

Our recipe of scaling CLIP to native worldwide data and training comprises three steps shown in Fig. \ref{fig:pipeline}: (1) constructing worldwide metadata, (2) implementing worldwide curation algorithm, and (3) building training framework for worldwide model. 
For generalizable recipe and findings, Meta CLIP~2 is designed to maximize overlapping with OpenAI CLIP and Meta CLIP, and only adopts necessary changes to learn from worldwide data.

\subsection{Revisit of Meta CLIP Algorithm}
\label{sec:Meta CLIP_algorithm}

We revisit the original Meta CLIP algorithm to illustrate how English-based CLIP data is curated with metadata constructed from human knowledge. 
The algorithm first constructs \textbf{metadata $\mathcal{M}$}, a list of high-quality visual concepts, from corpora written by human experts. $\mathcal{M}$ contains 500k entries, a combination and deduplication of entities from four high-quality sources: 1) all English WordNet Synsets, 2) Wikipedia English unigrams, and 3) bigrams, and 4) Wikipedia page titles. Then, the algorithm performs \textbf{substring matching}
on each alt-text (from a given image-text pair in the data pool $\mathcal{D}$) using metadata $\mathcal{M}$ to obtain a list \texttt{matched\_entry\_ids}.
\textbf{Global counting} is conducted to calculate the number of matches over $\mathcal{D}$ for each entry in $\mathcal{M}$
as \texttt{entry\_count}. Finally, the algorithm applies \textbf{balancing} to transform the raw image-text pair distribution into a distribution that is balanced for head and tail concepts and ready for training, by associating each pair with a sampling probability. Specifically, the count per entry is first converted into a probability of sampling each entry, \texttt{entry\_prob}, where tail entries (defined as \texttt{entry\_count} < $t$) have a probability set to 1, and all the other head entries have \texttt{$t$/entry\_count} as sampling probabilities. Each pair is then sampled based on probabilities of matched entries in its alt-text. Here, $t$ is a threshold to decide head vs. tail entries and set to 20k in OpenAI CLIP; Meta CLIP raised $t$ to 170k for scaling to billion English pairs.

\subsection{Worldwide Metadata}

We address the first challenge for worldwide scaling by constructing 
the missing metadata to cover the non-English world.
We maintain independent metadata per language since such design is intuitive (e.g., the same word ``mit'' has different meaning in English and Germany), has better performance (see ablation in Sec.~\ref{sec:ablation_metadata_curation_tokenizer}), and is flexible for adding and curating a new set of languages in future.

Our metadata is from the same four sources as OpenAI CLIP and Meta CLIP, but beyond English. Key changes are highlighted as follows. 
1) Multilingual WordNet: we include all synsets from 31 languages. 2) Wikipedia Unigrams and 3) Bigrams: we process unigram and bigram from Wikipedia dumps dated on May 2024, which include corpora in 329 languages. We clean the corpora into plain text with WikiExtractor~\citep{wikiextractor}. For most languages, we use space and punctuation to tokenize text into words, and then count unigrams and bigrams. For languages without space separation (e.g., some Asian languages), we use open-source tokenizers (see Table~\ref{tbl:tokenizer} in Appendix) developed by local communities to properly split text into words and meanwhile maintain the semantic integrity. 
4) Wikipedia Titles: we use page titles from 40 random dates of Wikipedia snapshots and rank these titles by click-through traffic for each language.

\subsection{Curation Algorithm}

Next, we scale curation to worldwide data language-by-language. The curation algorithm is detailed below and summarized in pseudo-code as Algorithm \ref{alg:code}. First, we conduct \textbf{language identification} (LID)~\citep{grave2018learning} to classify the language of the alt-text from an image-text pair, and choose language-specific metadata to match concepts. 
The sets of languages covered by LID and metadata sources (e.g., Wikipedia) are usually different, so we first establish a mapping between one language in LID to a unique set of languages in metadata entries. The languages in the metadata mapped to the same language in LID are merged into one group. This ends with a dictionary representation of metadata, \texttt{M}, where the keys are each language in LID and the values are the combined metadata of each group of languages. We also include a key ``other'' for metadata of languages that cannot be associated with any language in LID. Each
alt-text (\texttt{text}) in $\mathcal{D}$ is applied with LID for predicting its language (\texttt{text.lang}). After that, as in the Meta CLIP algorithm summarized in Sec.~\ref{sec:Meta CLIP_algorithm}, we run \textbf{substring matching} with metadata corresponding to predicted languages: \texttt{matched\_entry\_ids = substr\_match(text, M[text.lang])}, and aggregate \textbf{global count}, the number of matches of each entry, in \texttt{entry\_counts}.

With counts calculated, we \textbf{balance} occurrence of concepts across pairs. In data curation for English CLIP described above, a threshold $t$ is designed to limit the \emph{matches per metadata entry}, where entries with matches fewer than $t$ are defined as tail entries (or concepts) and otherwise head. 
Image-text pairs from head concepts are downsampled by a sampling probability derived from $t$ to balance training data distribution.
Thus, $t$ depends on the size of raw data pool (e.g., a larger pool has higher counts for the same entry).
OpenAI CLIP sets $t$ to 20k for 400M pairs; Meta CLIP~\citep{xu2024demystifying} tunes $t$ to 170k for scaling the training dataset to 2.5B and keeping the same ratio, 6\% of matches from \emph{tail concepts}, that OpenAI CLIP leverages to obtain the 400M pairs.
For worldwide data, the data size and the counts of matches differ greatly across languages, so $t$ should be language-\emph{dependent}.
Applying a single threshold $t$ to all languages yields suboptimal performance, e.g., a larger $t$ for a language with fewer pairs may yield too many pairs of head concepts and dilutes tail concepts in the curated data (see Sec.~\ref{sec:ablation_metadata_curation_tokenizer}). 

To derive $t$ for each language, we leverage the \emph{invariance} assumption adopted in Meta CLIP algorithm design, the percentage of \emph{tail matches} (i.e., 6\%), and apply it across languages.
With this assumption, we determine $t$ in two steps. 
(1) \textbf{From} $t_\text{en}$ \textbf{to $p$}: we calculate the global tail proportion $p$ for all languages, based on matches of English tail entries decided by $t_\text{en}$.
(2) \textbf{From $p$ to} $t_\text{lang}$: for each non-English language, we reversely find the language-specific threshold $t_\text{lang}$ based on the calculated $p$ to ensure the same tail proportion across all languages.
Detailed implementation of these two steps is shown as the \texttt{t\_to\_p()} and \texttt{p\_to\_t()} functions in Algorithm~\ref{alg:code}. With $t_\text{lang}$, \texttt{entry\_counts} is converted to \texttt{entry\_probs} similarly as in Meta CLIP but for each language.

Putting everything computed together, 
Algorithm \ref{alg:code} takes raw image-text pairs $\mathcal{D}$, metadata \texttt{M}, and an arbitrary threshold for English $t_\text{en}$ as input, and outputs a curated dataset of balanced and diverse training pairs, $\mathcal{D}^\ast$, with three stages.
\textbf{Stage 1} performs language-specific substring matching for each alt-text, \texttt{text}, based on LID results and corresponding metadata, and obtains match counts, \texttt{entry\_counts}, for each language and entry.
\textbf{Stage 2} computes thresholds $t_\text{lang}$ from $t_\text{en}$.
\textbf{Stage 3} samples image-text pairs based on matched entries in \texttt{text} with probabilities \texttt{entry\_probs}. Pairs matched with tail entries are always selected (i.e., probability = 1.0); pairs with head entries have sampling probabilities $t_\text{lang}$ / \texttt{entry\_counts[lang]}.
Sampled pairs compose $\mathcal{D}^\ast$.

\begin{algorithm}[ht]
\caption{Pseudo-code of Meta CLIP~2 Curation Algorithm in Python/NumPy.}
\label{alg:code}
\definecolor{codeblue}{rgb}{0.25,0.5,0.5}
\lstset{
  backgroundcolor=\color{white},
  basicstyle=\fontsize{8pt}{9pt}\ttfamily\selectfont,
  columns=flexible,
  breaklines=true,
  captionpos=b,
  commentstyle=\fontsize{9pt}{9pt}\color{codeblue},
  keywordstyle=\fontsize{9pt}{9pt},
}
\centering
\begin{lstlisting}[language=python]
"""
Input: 
D(list) raw (image, text) pairs: each text is assigned with a language "text.lang" by LID;
M(dict) worldwide metadata: key->language code; value(list)->metadata for that language;
t_en(int) English threshold on counts of head/tail entry cutoff: OpenAI CLIP->20k, Meta CLIP->170k;

Output: 
D_star(list): curated image-text pairs;
"""

# helper functions to compute t for each language.
def t_to_p(t, entry_count):
    return entry_count[entry_count < t].sum() / entry_count.sum()

def p_to_t(p, entry_count):
    sorted_count = np.sort(entry_count)
    cumsum_count = np.cumsum(sorted_count)
    cumsum_prob = cumsum_count / sorted_count.sum()
    return sorted_count[(np.abs(cumsum_prob - p)).argmin()]

# Stage 1: sub-string matching.
entry_counts = {lang: np.zero(len(M[lang])) for lang in M}
for image, text in D:
    # call substr_match which returns matched entry ids.
    text.matched_entry_ids = substr_match(text, M[text.lang])
    entry_counts[text.lang][text.matched_entry_ids] += 1

# Stage 2: compute t for each langauge.
p = t_to_p(t_en, entry_counts["en"]); t = {}
for lang in entry_counts:
    t[lang] = p_to_t(p, entry_counts[lang])

# Stage 3: balancing via indepenent sampling per language.
entry_probs = {}
for lang in entry_counts:
    entry_counts[lang][entry_counts[lang] < t[lang]] = t[lang]
    entry_probs[lang] = t[lang] / entry_counts[lang]

D_star = []
for image, text in D:
    for entry_id in text.matched_entry_ids:
        if random.random() < entry_probs[text.lang][entry_id]:
            D_star.append((image, text))
            break
\end{lstlisting}
\end{algorithm}

\subsection{Training Framework}
\label{sec:scaling_recipe}

Adopting data prepared with worldwide curation in current CLIP training framework addresses the first challenge, but curse of multilinguality still exists as shown in Fig. \ref{fig:teaser}.
Thus, we further design the worldwide CLIP training framework.
To make our framework and findings generalizable to CLIP and its variants, our framework follows OpenAI/Meta CLIP's training setting and model architecture with three additions: (1) a multilingual text tokenizer, (2) scaling seen training pairs, and (3) 
study of minimal viable model capacity.
The first is required to support worldwide languages and discussed in Sec.~\ref{sec:ablation_metadata_curation_tokenizer} for various choices; details of the latter two are described below.

\paragraph{Scaling seen pairs.}
Expanding from an English-only dataset and distribution to worldwide naturally increases the number of available image-text pairs. Training CLIP for worldwide distribution with the same number of seen pairs as English CLIP downsamples English training pairs and harms English performance. Hence, we scale seen pairs proportionally to the growth of data size from non-English pairs, to ensure the amount of English seen pairs unchanged during the worldwide CLIP training. This is achieved by increasing the global training batch size, which encourages cross-lingual learning, and meanwhile keeping the other training hyperparameters unchanged. 
We choose a 2.3$\times$ scaling of global batch to reflect that English pairs constitute 44\% of our training data. We ablate other choices of global batch size in Sec.~\ref{sec:main_ablation}.

\paragraph{Minimal viable model capacity.}  
Lastly, we study the minimal model expressivity to enable learning on extra seen pairs and break the curse of multilinguality. As in Fig. \ref{fig:teaser}, we find that even a ViT-L/14 (largest model provided by OpenAI) suffers from the curse due to deficient capacity, and
ViT-H/14 is the inflection point to break the curse (strong performance improvement in both English and non-English tasks).

\section{Experiment}
\label{sec:experiment}

\subsection{Dataset and Training Setup}
Following Meta CLIP pipeline, we collect image-text pairs sourced from the Internet that are publicly available.
After LID, there are about 44\% of alt-texts are in English, which are on par with the scale of English-only data from Meta CLIP~\citep{xu2024demystifying}. 
For generalizable recipe and findings, we base our training setup on OpenAI CLIP's ViT-L/14 and Meta CLIP ViT-H/14, except changes necessary for enabling worldwide capability, as described in Sec.~\ref{sec:scaling_recipe} and ablated in later subsections. The full details can be found in Table \ref{tbl:hp} and Appendix~\ref{sec:training_setup}.

\subsection{Evaluation}
\label{sec:main}

We first present the main ablations of Meta CLIP~2 on a wide range of English and multilingual zero-shot transfer benchmarks, along with other multilingual CLIP baselines for comparison (Sec.~\ref{sec:main_ablation}); then 
we conduct a comprehensive ablation study on the variants of metadata, curation and tokenizer (Sec.~\ref{sec:ablation_metadata_curation_tokenizer}).
Lastly, we evaluate the embedding quality of Meta CLIP~2 on downstream tasks for culture diversity (Sec.~\ref{sec:cultural}).
Additionally, we conduct analysis on embedding alignment and uniformity~\citep{wang2020hypersphere} in Sec.~\ref{appx:align_uni}.

\subsubsection{Main Ablation}
\label{sec:main_ablation}
We first ablate the effects of scaling seen training pairs and minimal viable model capacity that break the curse of multilinguality, 
with the following two groups of 6 training runs. Two trainings are in ViT-L/14 on worldwide curated data and its English portion, where global batch size and seen pairs are set to 2.3$\times$ and 1.0$\times$ compared to OpenAI CLIP and Meta CLIP setting (i.e., 1.0$\times$ has 12.8B seen pairs, or 400M for 32 epoches as in OpenAI CLIP). Four runs are on ViT-H/14 with different subsets of curated data to demonstrate the effects of English data helping multilingual performance and vice versa. We denote each run based on subsets trained with and corresponding seen pairs: 1) Worldwide (2.3$\times$) with the full-fledged worldwide curated data; 2) Worldwide (1.0$\times$) with 1) downsampled; 3) English (1.0$\times$) with English portion of 1); 4) Non-English (1.3$\times$) with the non-English portion.

We adopt the following two groups of zero-shot transfer benchmarks: 1) \textit{English-only} benchmarks on \textbf{ImageNet (IN val)}~\citep{russakovsky2015imagenet}, 
\textbf{SLIP 26 tasks (SLIP 26 avg.)}~\citep{mu2021slip},
and \textbf{DataComp 37 tasks (DC 37 avg.)}~\citep{gadre2023datacomp};
2) \textit{multilingual} benchmarks on \textbf{Babel-ImageNet (Babel-IN)}~\citep{geigle2024babel} (averaged zero-shot classification on IN with classes and prompts translated into 280 languages), 
\textbf{XM3600}~\citep{thapliyal2022crossmodal} (multilingual text-to-image, T→I, and image-to-text, I→T, retrieval with an averaged  recall@1 on 36 languages),
\textbf{CVQA}~\citep{mogrovejo2024cvqa} (multilingual multi-choice visual question answering with English and local averaged answer accuracy),
\textbf{Flickr30k-200}~\citep{visheratin2023nllb} (Flickr30k test set translated into 200 languages), 
\textbf{XTD-10}~\citep{aggarwal2020towards} (multilingual image-text retrieval on MSCOCO~\citep{chen2015microsoft} averaged Recall@1 over 7 languages), and \textbf{XTD-200}~\citep{visheratin2023nllb} (XTD10 translated into 200 languages). The main ablation is shown in Table~\ref{tab:main_results}. We observe that Meta CLIP~2 on ViT-H/14 with worldwide data and scaled seen pairs consistently outperforms its counterparts English (1.0$\times$) and Non-English (1.3$\times$), on both English and multilingual tasks, effectively breaking the
``curse of multilinguality''. The curse still exists in non-scaled seen pairs, Worldwide (1.0$\times$) or smaller ViT-L/14 model even with Worldwide (2.3$\times$)).

\begin{table}[h]
\centering
\scalebox{0.58}{
\begin{tabular}{l|c|c|l|c|c|c|c|c|c|c|c|c}
\hline
& & & & \multicolumn{3}{c|}{\textbf{English Benchmarks}} & \multicolumn{6}{c}{\textbf{Multilingual Benchmarks}} \\
\cline{5-13}
\textbf{Model}  & \makecell{\textbf{ViT}\\\textbf{Size (Res.)}} & \textbf{Data} & \makecell{\textbf{Seen}\\\textbf{Pairs}} & 
\makecell{\textbf{IN}\\\textbf{val}} & \makecell{\small{\textbf{SLIP 26}}\\\textbf{avg.}} & \makecell{\small{\textbf{DC 37}}\\\textbf{avg.}} & \makecell{\textbf{Babel}\\\textbf{-IN}} & 

\makecell{\textbf{XM3600}\\T$\rightarrow$I I$\rightarrow$T} & 

\makecell{\textbf{CVQA}\\EN LOC} & 

\makecell{\small{\textbf{Flicker30k}}\\\small{\textbf{-200}}\\T$\rightarrow$I I$\rightarrow$T} & 

\makecell{\textbf{XTD-10}\\T$\rightarrow$I I$\rightarrow$T} & 

\makecell{\textbf{XTD-200}\\T$\rightarrow$I I$\rightarrow$T} \\
\hline
\g{\small{XLM-CLIP}\citep{ilharco_gabriel_2021_5143773}} & \g{H/14(224)} & \g{\small LAION-5B} & \g{32B (2.5$\times$)} & \g{77.0} & \g{69.4} & \g{65.5} & \g{34.0} & \g{50.4 /} \g{60.5} & \g{56.1 /} \g{48.2} & \g{43.2 /} \g{46.2} & \g{87.1 /} \g{88.4} & \g{42.5 /} \g{45.2} \\

\g{mSigLIP\citep{zhai2023sigmoid}} & \g{B/16(256)} & \g{WebLI(12B)} & \g{40B (3.0$\times$)} & \g{75.1} & \g{63.8} & \g{60.8} & \g{40.2} & \g{44.5 /} \g{56.6} & \g{51.8 /} \g{45.7} & \g{34.0 /} \g{36.0} & \g{80.8 /} \g{84.0} & \g{37.8 /} \g{40.6} \\

\g{mSigLIP\citep{zhai2023sigmoid}} & \g{\small SO400M(256)} & \g{WebLI(12B)} & \g{40B (3.0$\times$)} & \g{80.6} & \g{69.1} & \g{65.5} & \g{46.4} & \g{50.0 /} \g{62.8} & \g{56.8 /} \g{49.8} & \g{39.9 /} \g{42.0} & \g{85.6 /} \g{88.8} & \g{42.5 /} \g{45.2} \\
\g{SigLIP~2\citep{tschannen2025siglip}} & \g{\small SO400M(256)} & \g{WebLI(12B)} & \g{40B (3.0$\times$)}  & \g{83.2} & \g{73.7} & \g{69.4} & \g{40.8} & \g{48.2 /} \g{59.7} & \g{58.5 /} \g{49.0} & \g{36.6 /} \g{40.3} & \g{86.1 /} \g{87.6} & \g{40.3 /} \g{44.5}\\
\hline
\multirow{2}{*}{Meta CLIP\citep{xu2024demystifying}} & L/14(224) & English(2.5B) & 13B (1.0$\times$) & 79.2 & 69.8 & 65.6 & - & - \quad - & - \quad - & - \quad - & - \quad - & - \quad - \\
 & H/14(224) & English(2.5B) & 13B (1.0$\times$) & 80.5 & 72.4 & 66.5 & - & - \quad - & - \quad - & \g{- \quad -} & - \quad - & - \quad - \\
\hline

\multirow{2}{*}{Meta CLIP~2} & \multirow{2}{*}{L/14(224)} & English & 13B (1.0$\times$) & 79.5 & 69.5 & 66.0 & - & - \quad - & - \quad - & - \quad - & - \quad - & - \quad -\\

& & Worldwide & 29B (2.3$\times$) & 78.8 & 67.2 & 63.5 & 44.2 & 45.3 / 58.2 & 59.2 / 55.1 & 41.9 / 45.8 & 82.8 / 85.0 & 41.9 / 44.8\\
\hline
\multirow{4}{*}{Meta CLIP~2} & \multirow{4}{*}{H/14(224)} & English & 13B (1.0$\times$) & 80.4 & 72.6 & 68.7 & - & - \quad - & - \quad - & - \quad - & - \quad - & - \quad -\\

&  & Non-Eng. & 17B (1.3$\times$)& 71.4 & 63.1 & 61.7 & 49.9 & 46.9 / 59.9 & 59.8 / 56.8 & 47.5 / 50.5 & 83.2 / 85.7 & 46.6 / 49.2\\

&  & Worldwide & 13B (1.0$\times$)& 79.5 & 71.1 & 67.2 & 47.1 & 49.6 / 62.6 & 59.9 / 56.0 & 49.1 / 52.1 & 85.2 / 87.1 & 47.0 / 49.7\\
\rowcolor{lightpurple}
\cellcolor{white} & \cellcolor{white} & Worldwide & 29B (2.3$\times$) & 81.3 & 74.5 & 69.6 & 50.2 & 51.5 / 64.3 & 61.5 / 57.4 & 50.9 / 53.2 & 86.1 / 87.5 & 48.9 / 51.0\\
\hline
\end{tabular}

}
\vspace{1.5mm}
\caption{Main ablation: Meta CLIP~2 breaks the curse of multilinguality when adopting ViT-H/14, with seen pairs scaled (2.3$\times$) proportional to the added non-English data. Meta CLIP~2 outperforms mSigLIP with fewer seen pairs (72\%), lower resolution (224px vs. 256px), and comparable architectures (H/14 vs. SO400M). 
We \g{grey out} baselines those are SoTA-aiming systems with confounding factors. Here, numbers of seen pairs are rounded to the nearest integer (e.g., 12.8B->13B).
}
\label{tab:main_results}
\end{table}

Although SoTA is non-goal for Meta CLIP~2, its full recipe
demonstrates strong performance
with fewer seen pairs (72\% of SigLIP series) and lower resolution (224px vs mSigLIP's 256). Meta CLIP~2 surpasses mSigLIP on IN, SLIP 26, and DC 37, and the recent SigLIP 2 on later two. More significantly, Meta CLIP~2 sets many SoTA multilingual benchmarks, e.g., Babel-IN (+3.8\%), XM3600 (+1.1\%/+1.5\%), CVQA (+3\%/+7.6\%), Flicker-30k-200 (+7.7\%/+7\%), and XTD-200 (+6.4\%/+5.8\%). SigLIP~2 prioritizes English (90\% of its training data in English), while it is worse than mSigLIP on multilingual tasks and Meta CLIP~2 on most English benchmarks except IN.

\begin{table}[h]
\centering
\small
\scalebox{0.88}{
\begin{tabular}{l|l|l|c|c|c c|c c}
\hline

\textbf{Ablation Steps} & \textbf{Metadata} & \textbf{Alt-texts} & \textbf{IN} & \textbf{Babel-IN} & \multicolumn{2}{c|}{\makecell{\textbf{XM3600}\\T$\rightarrow$I I$\rightarrow$T} } & \multicolumn{2}{c}{\makecell{\textbf{CVQA}\\EN LOCAL} }\\
\hline
1: English CLIP & English & English & \bf 67.5 & - & - & - & - & - \\
2: remove English filter & English & all, in 1 set & 66.9 & - & - & - & - & - \\

3: no language isolation & all, in 1 set & all, in 1 set & 62.1 & 31.2 & 37.8 & 49.7 & 49.8 & 45.8\\

4: language isolation with $t_\text{lang}=t_\text{en}$ & all, by lang. & all, by lang. & 61.1 & \bf 31.5 & 37.9 & 49.4 & 49.0 & 46.5\\

5: language specific $t_\text{lang}$ & all, by lang. & all, by lang. & 64.7 & \bf 31.5 & \bf 38.1 & \bf 50.0 & \bf 50.3 & \bf 46.6\\
\hline
\end{tabular}
}
\vspace{1.5mm}
\caption{Ablation study of metadata and alt-texts combination on ViT-B/32 using English 1.0$\times$ and Worldwide 1.0$\times$ with mT5 multilingual tokenizer. $t_\text{lang}$ is the count thresholds for each language and $t_\text{en}$ for English.}
\label{tbl:ablation_metadata_curation}
\end{table}
\vspace{-3.mm}

\subsubsection{Ablation on Metadata, Curation, and Tokenizer}
\label{sec:ablation_metadata_curation_tokenizer}
We further ablate the transition from metadata and curation focuses solely on English to their worldwide equivalents using the ViT-B/32 encoder for efficiency.
We evaluate zero-shot transfer on IN for English and Babel-IN, XM3600 and CVQA for multilingual. As in Table~\ref{tbl:ablation_metadata_curation}, starting from English-only CLIP, we first remove the English filter on alt-texts so that all alt-texts are curated by English metadata, resulting in 0.6\% drop on IN, indicating English isolation separating text or metadata by LID before matching is important. Then, we replace English metadata using all metadata merged without separation, yielding even worse English performance but start building up multilingual capability. Next, we isolate substring matching and curate alt-text language-by-language, with the same $t_\text{en}$ over all languages. This further lowers English performance since $t_\text{en}$ is too high for non-English and let head data dominate curation.
Lastly, we compute $t_\text{lang}$, to keep the same ratio of head-to-tail concepts for each language. This improves English and non-English performance, while curse of multilinguality remains unresolved in ViT-B/32 until the main ablation described above.

To minimize changes in model architecture, we only swap the English tokenizer for a multilingual one. Four popular tokenizers are studied on our zero-shot benchmarks. As shown in Table~\ref{tbl:tokenizer_ablation}, the XLM-V vocabulary yields the strongest performance in both the English and non-English world.

\begin{table}[h]
\centering
\scalebox{0.9}{
\begin{tabular}{l|c|c|c|c c|c c}
\hline

\textbf{Tokenizer} & \textbf{Vocab. Size} & \textbf{IN val} & \textbf{Babel-IN avg.} & \multicolumn{2}{c|}{\makecell{\textbf{XM3600}\\T$\rightarrow$I I$\rightarrow$T} } & \multicolumn{2}{c}{\makecell{\textbf{CVQA}\\EN LOCAL} } \\
\hline
mT5 (mSigLIP) \citep{xue2020mt5} & 250k & \bf 64.7 & 31.5 & 38.1 & 50.0 & 50.3 & 46.6 \\
Gemma (SigLIP~2) \citep{team2024gemma} & 256k & 63.7 & 26.1 & 36.1 & 47.8 & 48.3 & 44.0 \\
XLM-Roberta \citep{conneau2019unsupervised} & 250k & 64.0 & 31.1 & 38.0 & 49.8 & 49.8 & 46.1\\
XLM-V \citep{liang2023xlm} & 900k & \textbf{64.7} & \textbf{32.7} & \textbf{40.0} & \textbf{51.4} & \textbf{50.4} & \textbf{47.4} \\
\hline
\end{tabular}
}
\vspace{1.5mm}
\caption{Ablation study of various multilingual tokenizers with ViT-B/32 and Worldwide 1.0$\times$.}
\label{tbl:tokenizer_ablation}
\end{table}
\vspace{-3.mm}

\subsubsection{Cultural Diversity}
\label{sec:cultural}

\begin{figure}[h]
    \centering
    \includegraphics[width=\textwidth]{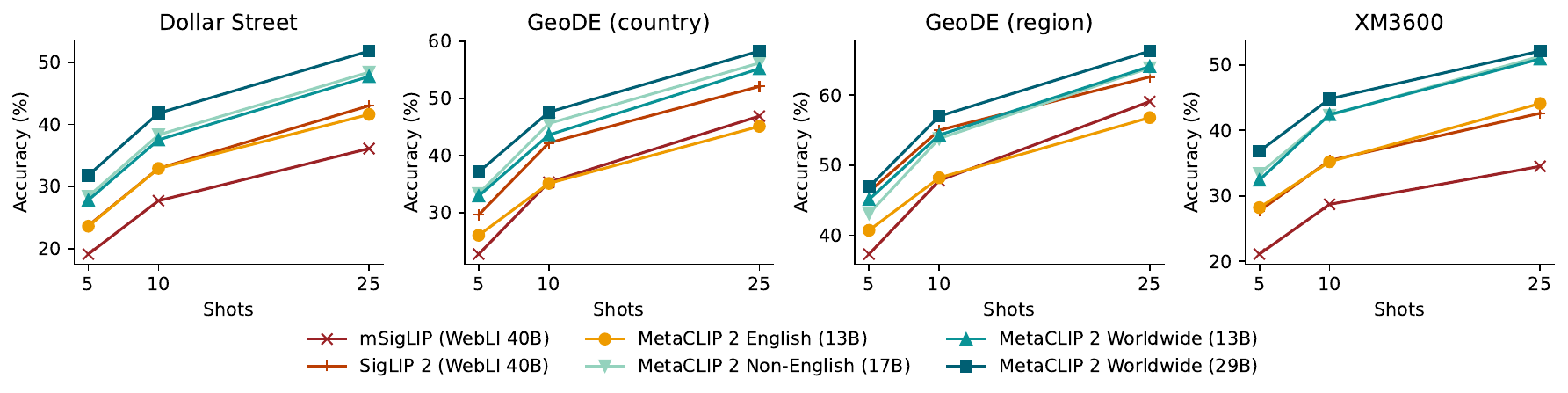}
    \caption{Few-shot geo-localization accuracy on cultural diversity benchmarks.
    }
    \label{fig:culture}
\end{figure}

Following protocols in \cite{pouget2024no} and \cite{wang2025scaling}, we perform zero-shot classification and few-shot geo-localization on a range of geographically diverse benchmarks. Specifically, we include zero-shot classification with Dollar Street~\citep{gaviria2022dollar}, GeoDE~\citep{ramaswamy2023geode}, and GLDv2~\citep{weyand2020google} in Table~\ref{tbl:culture}, and few-shot geo-localization~\citep{pouget2024no} on Dollar Street, GeoDE and XM3600 
in Fig.~\ref{fig:culture}. We find that only changing the training data distribution, from 13B \emph{English} to 13B \emph{worldwide} pairs, yields significantly better performance, and scaling to 29B \emph{worldwide} pairs improves further,
except for the on-par, probably saturated performance in GeoDE. Fig. \ref{fig:culture} shows similar trend for evaluating on few-shot geo-localization.
\vspace{-2.mm}

\begin{table}[h]
\centering
\scalebox{0.87}{
\begin{tabular}{l|l|l|c c c c}
\hline
\textbf{Model} & \textbf{Data} & \textbf{Seen Pairs} & \multicolumn{2}{c}{\makecell{\textbf{Dollar Street} \\ Top-1\quad Top-5}} & \textbf{GLDv2} & \textbf{GeoDE} \\
\hline
\g{mSigLIP~\citep{zhai2023sigmoid}} & \g{WebLI(12B)~\citep{chen2023pali}} & \g{40B (3.0$\times$)} & \g{36.0} & \g{62.5} & \g{45.3} & \g{94.5} \\
\g{SigLIP~2~\citep{tschannen2025siglip}} & \g{WebLI(12B)~\citep{chen2023pali}} & \g{40B (3.0$\times$)} & \g{36.7} & \g{61.9} & \g{48.5} & \g{95.2} \\
\hline
\multirow{4}{*}{Meta CLIP~2} & English & 13B (1.0$\times$) & 37.2 & 63.3 & 52.8 & 93.4 \\
& Non-English & 17B (1.3$\times$) & 35.7 & 61.3 & 68.6 & 91.7 \\
& Worldwide & 13B (1.0$\times$) & 37.2 & 63.7 & 65.8 & 94.3 \\
\rowcolor{lightpurple}
\cellcolor{white} & Worldwide & 29B (2.3$\times$) & 37.9 & 64.0 & 69.0 & 93.4 \\
\hline
\end{tabular}
}
\caption{Zero-shot classification accuracy on cultural diversity benchmarks. Meta CLIP~2 models are in ViT-H/14 and mSigLIP/SigLIP~2 are in ViT-SO400M. mSigLIP/SigLIP~2 are SoTA-aiming systems with many factors changed and thus greyed out.}

\label{tbl:culture}
\vspace{-3.5mm}
\end{table}

\subsubsection{Alignment and Uniformity}
\label{appx:align_uni}
\begin{figure}[h]
    \centering
    \includegraphics[width=0.8\textwidth]{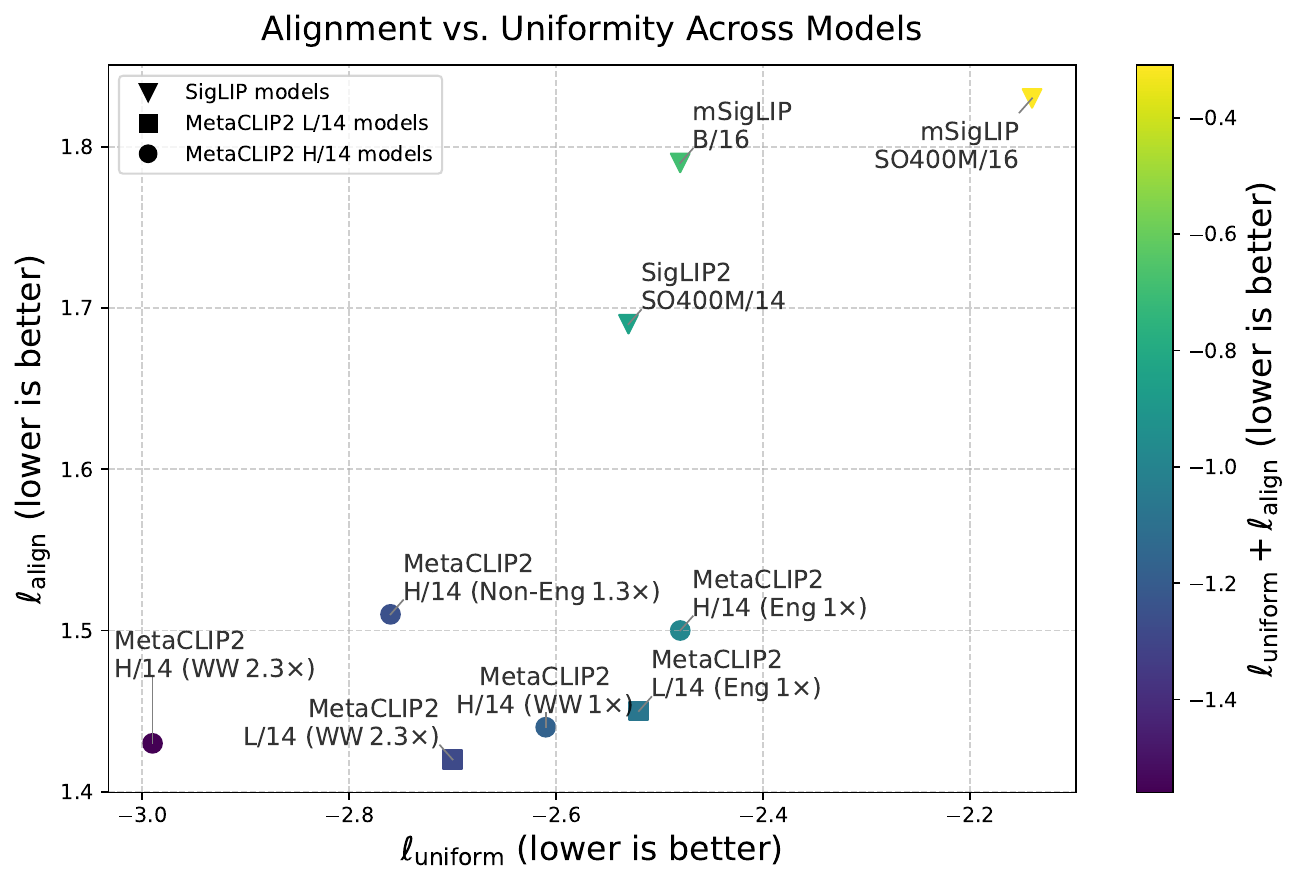}
    \caption{Alignment and uniformity scores~\citep{wang2020hypersphere} calculated on our collected 5k holdout data, WW indicates worldwide data.
    }
    \label{fig:alignment_uniformity}
\end{figure}

Following~\citep{wang2020hypersphere}, we further measure the embeddings quality across different CLIP models.
To avoid various unknown biases from different benchmarks, we use 5k holdout image-text pairs not used in our training and report alignment and uniformity scores, where alignment measures the relevance of an image and a text and uniformity measures how images distributed in vision encoder's embedding space. Note that we have no control on whether these 5k pairs are leaked in other baselines.
From Fig. \ref{fig:alignment_uniformity}, we can see that Meta CLIP~2 exhibits good scores in both alignment and uniformity (lower is better), whereas mSigLIP or SigLIP~2 may have non-trivial bias on our collected holdout data.

\section{Conclusion}
\label{conclusion}

We present Meta CLIP~2, the first CLIP trained with worldwide image-text pairs from scratch. Existing CLIP training pipelines, designed primarily for English, cannot straightforwardly generalize to a worldwide setting without incurring an English performance degradation due to lack of curation for worldwide data or the ``curse of multilinguality''. Our careful study suggests that the curse can be broken by scaling metadata, curation, and training capacity, where English and non-English world benefit each other. Specifically, Meta CLIP~2 (ViT-H/14) surpasses its English-only counterpart on zero-shot IN (80.5\% → 81.3\%) and sets new SoTA on multilingual benchmarks such as XM3600, Babel-IN and CVQA with one single model. We envision our findings along with the fully open-sourced metadata, curation and training code encourage the community to move beyond English-centric CLIP and embrace the worldwide multimodal web.

\section*{Acknowledgments}
We thank Thao Nguyen and Bernie Huang for insightful discussion, Wei Cheng and Guan Pang for evaluation on downstream use cases.

\clearpage
\newpage
\bibliographystyle{assets/plainnat}
\bibliography{paper}

\clearpage
\newpage
\beginappendix

\section{Implementation Details for Metadata and Curation} 
\label{appx:details}

\subsection{Unigram and Bigram Tokenizer for Special Languages}
\label{appx:tokenizers}

Most modern languages around the world adopt writing systems that use ``spaces'' to separate words, except for some of the Asian languages, known as ``scriptio continua''\citep{scriptio_continua}. We find several open source tokenizers for many of these languages developed by local communities, as shown in the Table~\ref{tbl:tokenizer}, in order to properly split text into words while preserve semantic integrity. Note these tokenizers are only used to process Wikipedia dump labeled with the listed wiki codes (e.g., not on alt-texts' substring matching).

\begin{table}[h]
\centering
\scalebox{0.85}{
\begin{tabular}{c|c}
\toprule
\textbf{Wiki Code} & \textbf{Tokenizer Name} \\
\midrule
bo,dz & Tibetan Tokenizer\\
ja,ryu  & Japanese Tokenizer\\
km  & Khmer Tokenizer\\
lo & Lao Tokenizer\\
my & Myanmar Tokenizer\\
th & Thai Tokenizer\\
zh,zh\_classical,zh\_yue & Chinese Tokenizer\\
\bottomrule
\end{tabular}
}
\vspace{1.5mm}
\caption{Tokenizers for special languages.}
\label{tbl:tokenizer}
\end{table}

\subsection{Scaling Curation} 
\label{sec:impl_curation}
Worldwide scaling of data curation significantly increases time and space complexity due to store metadata across hundreds of languages. To efficiently handle this complexity, we leverage several efficient implementation to enable worldwide curation (that English only curation does not have):

\begin{itemize}
    \item \textbf{Efficient String Matching}: We adopt the Aho-Corasick algorithm~\footnote{\href{https://en.wikipedia.org/wiki/Aho-Corasick_algorithm}{\tt https://en.wikipedia.org/wiki/Aho-Corasick\_algorithm}}\footnote{\url{https://pypi.org/project/pyahocorasick}}, which utilizes prefix trees (tries), for rapid substring matching. The matching speed is about 2k times faster than Meta CLIP's brute-force implementation, enabling matching with million-scale metadata.
    \item \textbf{Lazy Metadata Loading}: We pre-build and store the metadata into an Aho-Corasick automaton for each language separately, loading these automaton dynamically and only when encountering a new language for alt-text during processing, thereby minimizing the total number of languages encountered for each shard of data and saving re-compiling time for automation on a new shard.
    \item \textbf{Memory Management for Probabilities}: To address memory constraints during sampling for balancing, we utilize memory-mapped file loading (mmap) to efficiently access counts per entry across all languages, preventing out-of-memory errors caused by loading all the counts from different languages.
\end{itemize}

These implementation choices ensure the worldwide data curation is computationally feasible and scalable to billions of image-text pairs from hundreds of languages. 

\paragraph{Mitigation and Benchmark Deduplication} We run a state-of-the-art safety classifier to remove NSFW contents (e.g., adult, sexual, violence) from training data. We also apply face detector to remove human biometrics and personally identifiable information from data. To avoid benchmark leakage, we remove any overlap with ImageNet evaluation sets by performing deduplication using 64-bit hashes. These hashes are generated by applying random projection to feature embeddings from a similarity search model, reducing them to 64 dimensions followed by sign-based quantization.

\section{Training Setup}
\label{sec:training_setup}
To remove confounding factors and generalize our findings, we follow OpenAI CLIP and Meta CLIP training setup with changes for worldwide scaling, detailed in Table~\ref{tbl:hp}.  
Our data curation algorithm is running in parallel with 800 jobs (each job has 40GB CPU memory) and it takes 1 hour to substring match and count for all alt-text pairs.

\begin{table}[h]
\centering
\scalebox{0.8}{
    \setlength\tabcolsep{3.0pt}
    \begin{tabular}{l| c | c}
    \toprule 
    Hyperparameter & OpenAI CLIP / Meta CLIP & Meta CLIP~2\\ 
    \midrule
        Activation Function & QuickGELU & QuickGELU \\
        Seen Pairs & 12.8B & 29B (2.3$\times$) \\
        Batch Size & 32768 & 75366 (2.3$\times$) \\
        Learning Rate & 4.0e-4 (L/14, H/14) & 4.0e-4 (H/14) \\
        Warm-up & 2k & 2k \\
    \bottomrule
    \end{tabular}
}

\caption{Hyperparameters of OpenAI CLIP / Meta CLIP vs Meta CLIP~2.}
\label{tbl:hp}
\end{table}

\section{Limitation on Benchmark}
\label{sec:lim_bench}

High-quality benchmarks are essential for researchers to understand the efficacy of proposed changes. After decades of meticulous efforts, the community has established reliable and diverse datasets to enable research advancement in vision and multimodal areas~\citep{deng2009imagenet,russakovsky2015imagenet,radford2021learning}. However, these datasets consist mainly of content scraped from North America and Western Europe (NA+EU) and focus on English~\citep{shankar2017no,de2019does}. It is a long and resource-intensive endeavor to build similar benchmarks for unbiased and comprehensive evaluation of worldwide data and resulting representations, for the world outside NA+EU or English-speaking community, due to the complexity of covering diverse concepts across geo-locations, cultures, and languages. XM3600~\citep{thapliyal2022crossmodal} aims to build geographically diverse datasets by selecting images from Open Images Dataset~\citep{kuznetsova2020open} based on metadata of GPS coordinates, but later research~\citep{pouget2024no} suggests Open Images Dataset is biased towards Western images or specific activities (e.g., tourism). 
GeoDE~\citep{ramaswamy2023geode} recruits human workers on crowdsourcing platform to collect geographically diverse images for predefined object classes. Crowdsourcing is an economic way to collect human annotations, but the demographic background and proficiency of the workers are not guaranteed, nor is the quality of the collected data.
Few efforts such as CVQA~\citep{mogrovejo2024cvqa}
attempt to scale annotation and control quality simultaneously 
by utilizing experts in machine learning community or existing materials as seeds. These efforts offer relatively unbiased evaluation with reasonable coverage in capabilities (e.g., cultural diversity, multimodal problem solving for exam questions across countries) of interests. We believe benchmarks of similar quality but built for evaluating more general and comprehensive capabilities will reveal the true potential of worldwide data and resulting representations developed in this work.

\end{document}